\documentclass[journal,comsoc]{IEEEtran}
\usepackage[T1]{fontenc}% optional T1 font encoding
\usepackage{cite}

\ifCLASSINFOpdf
   \usepackage[pdftex]{graphicx}
\else
   \usepackage[dvips]{graphicx}
\fi

\usepackage{amsmath}
\usepackage[cmintegrals]{newtxmath}

\usepackage[numbers]{natbib}
\usepackage{booktabs}
\usepackage{multirow}
\usepackage{multicol}
\usepackage{amsmath}
\usepackage{algpseudocode}
\usepackage{mathtools}
\usepackage[linesnumbered,ruled,vlined]{algorithm2e}

\mathchardef\mhyphen="2D

\begin{document}
\title{Continuous Cognitive Coverage for Autonomous Robots via Event-Dependent Cognitive Treatment and Learning}

\author{Hong~Su
% <-this % stops a space
\IEEEcompsocitemizethanks{\IEEEcompsocthanksitem H. Su is with the School of Computer Science, Chengdu University of Information Technology, Chengdu, China.\\
 E-mail: suguest@126.com. \\
\protect\\
}% <-this % stops an unwanted space
\thanks{}}

\markboth{Journal of \LaTeX\ Class Files,~Vol.~14, No.~8, August~2015}%
{Shell \MakeLowercase{\textit{et al.}}: Bare Demo of IEEEtran.cls for IEEE Communications Society Journals}
% * Note that you probably will NOT want to include the author's *
% * name in the headers of peer review papers.                   *
% You can use \IFCLASSOPTIONpeerreview for conditional compilation here if

\maketitle

\begin{abstract}
Autonomous robots continuously encounter objects, changes, and situations, and every event admitted into cognition should receive an appropriate cognitive treatment rather than remain untreated until an explicit task requires attention. However, existing task-driven, reactive, or fixed-reasoning approaches generally process only selected events or apply predefined reasoning procedures, making it difficult to provide continuous cognitive coverage with differentiated treatment. This paper proposes a continuous cognitive coverage framework in which every cognitively admitted event is assigned an event-dependent cognitive treatment according to its state, context, and history. Different events may therefore invoke description, memory, risk prediction, planning, diagnosis, analogy, or other learned treatments. Familiar events can be processed automatically by learned mechanisms, whereas unfamiliar or uncertain events invoke explicit deliberation or fallback reasoning. Multiple cognitive processes can be suspended, resumed, and interleaved so that cognitive processing continues as new events arrive or existing events await evidence. Validated experiences are continuously learned to automate, refine, and revise event-specific treatments. Experiments achieve 96.76\% structured treatment accuracy with 93.66\% automatic processing, 92.64\% cognitive coverage under bursty-delayed workloads, and 79.53\% continual-learning joint accuracy, with novel-event reuse reaching 100\% automatic processing.
\end{abstract}


\IEEEpeerreviewmaketitle

\section{Introduction}
\label{sec:introduction}

Autonomous robots operate in environments where objects, state changes, human
activities, and unexpected situations continuously appear while explicit tasks are
being executed.  Human cognition does not normally wait for a formally assigned task
before processing such events: encountered situations are noticed, interpreted, and
handled in different ways depending on their meaning and context.  A familiar and
harmless object may require only lightweight recognition, whereas an unexpectedly
moved object may trigger memory comparison, an unstable object may require risk
prediction, and an abnormal device may require diagnosis.  This suggests that
autonomous cognition should provide not only reasoning ability, but also
\emph{continuous cognitive coverage}: every event admitted into cognition should
receive an appropriate cognitive treatment.

Current robot-learning and reasoning systems only partially support this requirement.
Many approaches are task-centered, activating planning or reasoning mainly when an
explicit goal, query, or failure demands it.  Other methods employ a fixed reasoning
pipeline or a predefined mapping from event categories to processing procedures.
Such designs can be effective for a known task, but they do not explicitly address
three related questions: whether all cognitively admitted events remain covered while
the robot continues to operate, whether different events and contexts can receive
different forms of cognitive treatment, and whether these treatments can themselves
be continuously learned and revised.  In particular, applying an expensive reasoning
model to every event is unnecessary, while treating all familiar-looking events
automatically can also be unsafe when their context or outcome changes.

This paper proposes a \emph{continuous cognitive coverage} framework with
event-dependent cognitive treatment and learning.  Each cognitively admitted event is
first registered in a semantic representation and is then assigned an appropriate
treatment according to the event, its context, and relevant cognitive history.
Treatments may include description, memory, risk prediction, planning, diagnosis,
analogy, or other learned cognitive procedures.  The framework separately determines
\emph{what treatment} should be used and \emph{how it should be executed}.  Familiar
events can therefore be handled by learned automatic mechanisms, whereas uncertain,
unfamiliar, or changed events can invoke explicit deliberation or fallback reasoning.

Cognitive processing is maintained as a continuing stream rather than as an isolated
reasoning episode.  Multiple cognitive processes may coexist: one process can wait for
delayed evidence, another can be temporarily suspended, and a newly arriving critical
event can be processed before an earlier process is resumed.  Consequently, the end
or suspension of one cognitive process does not imply that cognition as a whole has
stopped.  This mechanism is intended to preserve cognitive coverage when events arrive
asynchronously or overlap in time.

The framework also treats learning as continuous and event-dependent.  Validated
experiences are used not only to learn the eventual result of an event, but also to
learn which cognitive treatment is appropriate for that type of event.  A treatment
that initially requires explicit reasoning can subsequently be reused automatically
after sufficient validation.  Conversely, if delayed evidence shows that a previously
reliable automatic treatment has become incorrect, the corresponding cognition can be
reopened, revised, and learned again.  Automatic processing is therefore viewed as a
learned implementation of cognition rather than the absence of cognition.

The main contributions of this paper are as follows:
\begin{itemize}
    \item We formulate \emph{continuous cognitive coverage}, requiring every
    cognitively admitted event to receive a cognitive treatment while allowing
    different events and contexts to invoke different treatments.

    \item We introduce a continuous process-management mechanism in which cognitive
    processes can be created, suspended, resumed, interleaved, and completed, enabling
    the robot to preserve cognitive coverage under concurrent and delayed events.

    \item We develop a continuous event-dependent learning mechanism that learns
    event-to-treatment mappings, converts validated treatments into automatic
    processing, and revises previously learned cognition when later outcomes
    contradict it.

\end{itemize}

The remainder of this paper is organized as follows. Section~\ref{sec:related_work} describes related work. Section~\ref{sec:model} introduces the continuous cognitive coverage model and its event-dependent treatment, continuous processing, and learning mechanisms.  Section~\ref{sec:verification} evaluates these mechanisms experimentally, followed by the conclusion.

\section{Related Work}
\label{sec:related_work}

\subsection{Cognitive Architectures and Continuous Cognitive Processing}

Classical cognitive architectures provide an important foundation for modeling
cognition as an ongoing process rather than as an isolated input--output function.
The LIDA architecture, for example, organizes cognition around repeated cognitive
cycles that integrate perception, attention, memory, action selection, and learning
\cite{franklin2012lida}.  Related psychologically oriented architectures such as
ACT-R, Soar, and CLARION also attempt to integrate multiple cognitive functions
within a unified computational framework, and several of them include both explicit
and implicit learning mechanisms \cite{helie2014autonomous}.  CLARION is particularly
relevant because it distinguishes explicit and implicit representations and supports
interaction between them during skill learning \cite{sun2001implicit}.  These systems
show that cognition can be modeled as a continuing interaction among specialized
processes rather than as a single monolithic reasoning routine.

The main strength of cognitive architectures is their broad treatment of cognition.
They explicitly represent internal processes, memory, attention, and learning, and
therefore provide a richer account than task-specific robot controllers.  However,
their objectives differ from the one considered here.  A cognitive cycle specifies
how cognition repeatedly proceeds, but does not by itself require that every
cognitively admitted event be assigned an explicit \emph{event-dependent cognitive
treatment}.  Similarly, explicit/implicit learning in architectures such as CLARION
primarily concerns the relationship between representational levels and skill
learning; it does not directly formulate the problem of learning which type of
cognitive treatment---for example memory, diagnosis, planning, or risk
prediction---should be applied to each encountered event.  Moreover, many cognitive
architectures rely on substantial predefined architectural structure and knowledge,
which can make adaptation to open-ended robotic event classes difficult
\cite{helie2014autonomous}.

Our work builds on the idea of continuous cognitive activity but focuses on a
different abstraction.  Instead of defining one global cognitive cycle, we treat
each cognitively admitted event as an object that must obtain cognitive coverage,
and we explicitly learn the mapping from the event, context, and history to an
appropriate treatment.  Multiple event-specific cognitive processes can coexist,
wait for evidence, be suspended, and later resume.  Thus, the proposed framework
complements classical cognitive architectures by making \emph{coverage} and
\emph{event-dependent treatment selection} first-class mechanisms.  It also connects
these mechanisms to learned automatic processing, so continuous cognition does not
require every event to execute the same expensive deliberative cycle.

\subsection{LLM-Based Embodied Reasoning and Robot Planning}

Recent work has shown that large language and multimodal models can substantially
improve high-level robotic reasoning.  SayCan grounds language-model proposals using
the affordances of pretrained robot skills, allowing semantic knowledge from an LLM
to be constrained by what a robot can actually execute \cite{ichter2023saycan}.
Inner Monologue further introduces closed-loop language feedback from scene
descriptions, success detectors, and human interaction so that an LLM can revise
plans during execution \cite{huang2023inner}.  PaLM-E incorporates continuous
sensor and visual information directly into an embodied multimodal language model
and demonstrates transfer across several embodied reasoning tasks
\cite{driess2023palme}.  RT-2 goes further toward end-to-end embodied control by
mapping vision and language to action tokens and transferring web-scale semantic
knowledge into robotic policies \cite{zitkovich2023rt2}.  More generally, ReAct
demonstrates the benefit of interleaving reasoning traces with actions and
environment interaction \cite{yao2023react}.

These approaches provide powerful semantic reasoning and task generalization, and
several explicitly use environmental feedback to correct a current plan.  Their
primary unit of reasoning, however, is usually a user instruction, task, planning
episode, or action-selection problem.  They do not normally address cognitive
coverage over all events that become relevant to an autonomous agent while another
task is already being executed.  Calling a general LLM for every admitted event would
also be computationally wasteful, whereas restricting reasoning to explicit tasks can
leave non-task events cognitively unattended.  In addition, embodied foundation
models usually learn a direct relation from observations and instructions to plans or
actions; they do not explicitly separate \emph{which cognitive treatment should be
used} from \emph{how that treatment should be executed}.

The proposed framework therefore uses LLM reasoning in a more selective role.
Explicit deliberation is one execution mode rather than the entire cognitive
architecture.  The robot first decides whether an event calls for description,
memory, planning, diagnosis, risk prediction, analogy, or another learned treatment,
and only uncertain or unfamiliar cases need expensive deliberation.  Familiar cases
can reuse a learned automatic processor.  This distinction also differs from
closed-loop replanning: a delayed diagnostic process may be suspended while unrelated
events continue to receive cognitive treatment, and it may later resume when evidence
arrives.  Consequently, our objective is not to replace embodied LLM planning, but to
provide a continuous event-management and treatment-selection layer around both
learned automatic mechanisms and explicit LLM reasoning.

\subsection{Continual Learning, Reflection, and Automatization of Cognition}

Continual and lifelong learning address the complementary problem of allowing an
autonomous system to improve while its data distribution and experience change over
time.  In robotics, continual learning is motivated by the need to acquire new
knowledge and skills from sequential experience without assuming that all training
data are available at once \cite{lesort2020continual}.  LLM-based embodied agents
have also explored persistent improvement without conventional offline retraining.
Voyager, for example, continuously explores an environment, builds an expanding
library of executable skills, and reuses those skills in later situations
\cite{wang2024voyager}.  Reflexion improves language-agent behavior by storing
linguistic reflections derived from feedback and using them in subsequent trials
\cite{shinn2023reflexion}.  These methods demonstrate the value of accumulating
experience, memory, and reusable competence over long interactions.

A related line of work studies how expensive deliberative reasoning can be converted
into faster processing.  Classical dual-representation models already investigated
interactions between explicit and implicit skill knowledge \cite{sun2001implicit}.
More recently, System-2-to-System-1 distillation has shown that outputs produced by
more expensive reasoning procedures can be compiled into faster direct model
responses, reducing inference cost while retaining some of the benefits of
deliberation \cite{yu2024distilling}.  This direction is closely related to the
automatic-reuse component of our framework.  Nevertheless, distillation normally
focuses on reproducing higher-quality outputs, whereas general continual-learning
methods focus on retaining or acquiring task competence.  Neither objective by itself
specifies that the learner should acquire an explicit event-to-cognitive-treatment
mapping or reopen previously automated cognition when a later outcome invalidates it.

Our work combines these ideas at the level of cognitive treatment.  Validated
experience is used to learn both \emph{what result is appropriate} and
\emph{what kind of cognitive process should handle the event}.  A treatment first
obtained through explicit reasoning can later be executed automatically, but the
automatic mechanism remains subject to delayed validation.  If the environment
changes and later evidence contradicts the learned result, confidence in the
automatic treatment can decrease and explicit cognition can be re-engaged before the
experience is learned again.  Thus, the contribution is not simply continual task
learning, reflection memory, or reasoning distillation.  It is a continuous loop in
which event-dependent cognitive treatments are acquired, automated, monitored, and
revised while cognitive coverage is maintained over the ongoing event stream.

\section{Continuous Cognitive Coverage with Event-Dependent Treatment and Learning}
\label{sec:model}

An autonomous robot continuously encounters objects, state changes, human activities,
environmental conditions, and other events while executing its tasks.  We argue that
cognition should not be activated only when an explicit task requires reasoning.
Instead, every event that is admitted into the robot's cognitive scope should receive
an appropriate cognitive treatment.  The treatment need not always be expensive
deliberation: a familiar event may be handled by a learned automatic mechanism,
whereas an unfamiliar, uncertain, or abnormal event may require explicit reasoning.
Different events may therefore receive different cognitive treatments, while the
overall cognitive process continues as events arrive, wait for evidence, complete,
or are revisited.

The proposed model is built around three related properties.  First,
\emph{cognitive coverage} requires that every cognitively admitted event be assigned
a treatment rather than being left cognitively unattended.  Second,
\emph{event-dependent cognitive treatment} allows different events, contexts, and
histories to invoke different ways of thinking.  Third, both cognitive processing
and the learning of these treatments are continuous, so validated experience can
progressively convert repeated deliberative treatment into automatic processing and
can revise previously learned treatment when later outcomes contradict it.

\subsection{Cognitive Coverage and Event-Dependent Treatment}
\label{subsec:cognitive_coverage}

Let $o_t$ denote the robot's observation at time $t$, and let $c_t$ denote the
current context, including task state, spatial relations, recent history, and other
information relevant to interpretation.  Cognitive registration transforms the
observation into an event representation

\begin{equation}
e_t = D(o_t,c_t),
\label{eq:cognitive_registration}
\end{equation}

where $D(\cdot)$ does not necessarily produce a natural-language description.  It may
instead produce a semantic feature representation, object--relation structure, scene
description, or another representation suitable for subsequent cognitive processing.
This registration step distinguishes a \emph{cognitively admitted event} from raw
sensory input.  The proposed framework does not require every low-level sensor sample
to trigger explicit cognition; rather, every event admitted by $D(\cdot)$ must be
cognitively treated.

Let $\mathcal{E}_t$ denote the set of events currently admitted into cognitive scope.
For each $e_i\in\mathcal{E}_t$, the robot assigns a cognitive treatment
$m_i$ from a treatment repertoire

\begin{equation}
\mathcal{M}_t =
\{M_1,M_2,\ldots,M_{K_t}\},
\label{eq:treatment_set}
\end{equation}

which may include description, memory recall, risk prediction, planning, diagnosis,
comparison, causal analysis, analogy, or other learned treatments.  Cognitive
coverage requires

\begin{equation}
\forall e_i\in\mathcal{E}_t,\qquad
\exists\, m_i\in\mathcal{M}_t
\ \text{such that}\
\mathrm{Covered}(e_i,m_i,t)=1 .
\label{eq:cognitive_coverage}
\end{equation}

Here, ``covered'' does not mean that the event must have completed all reasoning at
time $t$.  An event is cognitively covered if it has been registered and assigned an
appropriate cognitive process that is active, automatically handled, temporarily
suspended while retaining its state, or waiting for later evidence.

The appropriate treatment depends on the event rather than only on its object
category.  The selector is written as

\begin{equation}
(m_t^{*},z_t)
=
S(e_t,c_t,H_t),
\qquad
z_t\in
\{\mathrm{auto},\mathrm{delib},\mathrm{fallback}\},
\label{eq:treatment_selector}
\end{equation}

where $H_t$ denotes relevant validated cognitive history.  The variable $m_t^{*}$
specifies \emph{what kind of cognitive treatment} should be applied, whereas $z_t$
specifies \emph{how that treatment is executed}.  In the automatic mode, a learned
processor handles a familiar event without invoking explicit deliberation.  In the
deliberative mode, explicit reasoning is used when the event is difficult or uncertain.
In the fallback mode, a general or analogous treatment is provisionally applied when
no sufficiently reliable event-specific mechanism is available.

This separation is important because unfamiliarity is not itself a cognitive
treatment.  For example, an unfamiliar overheating device may still require diagnosis
or risk prediction; ``fallback'' only indicates that the treatment must currently be
obtained through a more general reasoning mechanism.  Likewise, the same object can
receive different treatments in different contexts.  A cup safely located at the
center of a table may require only description, the same cup near the table edge may
require risk prediction, and an unexpectedly moved cup may require memory comparison.
Thus, the framework learns and applies the mapping from an event together with its
context and history to an appropriate cognitive treatment.

\subsection{Continuous Cognitive Processing}
\label{subsec:continuous_processing}

Cognitive coverage must be maintained over time rather than only during isolated
task-level reasoning episodes.  The robot therefore maintains a dynamic set of
cognitive process instances

\begin{equation}
\mathcal{P}_t=
\{P_1^t,P_2^t,\ldots,P_{N_t}^t\},
\label{eq:cognitive_process_set}
\end{equation}

where each process corresponds to a cognitively admitted event and $N_t$ is not fixed.
A process may be newly created, active, waiting for evidence, suspended, resumed, or
completed.  When one process waits for delayed information, the robot can continue
processing other admitted events instead of blocking the entire cognitive stream.
Likewise, a newly arriving critical event can temporarily preempt a lower-priority
process without discarding the latter's cognitive state.

For example, suppose $P_1$ is diagnosing an abnormal device but cannot finish until
a later sensor reading becomes available.  While $P_1$ is suspended, a blocked route
can create $P_2$ for planning and an unstable object can create $P_3$ for risk
prediction.  When the missing evidence arrives, $P_1$ is resumed.  The resulting
sequence may therefore be

\begin{equation}
P_1^{(1)}
\rightarrow
P_2
\rightarrow
P_3
\rightarrow
P_1^{(2)},
\label{eq:interleaved_cognition}
\end{equation}

rather than forcing $P_1$ to finish before any subsequent event can be cognitively
treated.

Continuous cognition therefore does not mean one indefinitely long LLM call or one
reasoning chain that never terminates.  It means that the robot continuously maintains
cognitive coverage over the stream of admitted events.  Individual cognitive processes
may start and finish, but cognition as a system-level activity persists because new,
unfinished, suspended, and resumed processes coexist over time.

\subsection{Continuous Learning, Automatic Reuse, and Revision}
\label{subsec:continuous_learning}

Cognitive processing is continuous, and the learning of cognitive treatments is also
continuous.  After an event has been treated and later validated by an observed
outcome or additional evidence, the experience becomes learning material.  The robot
maintains

\begin{equation}
\mathcal{D}_t=
\{(e_i,m_i,r_i,y_i)\}_{i=1}^{n_t},
\label{eq:learning_material}
\end{equation}

where $e_i$ is the registered event, $m_i$ is the cognitive treatment applied to the
event, $r_i$ is the treatment result, and $y_i$ is the later validation signal or
observed outcome.  The dataset grows as new validated experiences become available.

A learned cognitive processor $F_{\theta}$ is trained not only to reproduce an
eventual result but also to learn which cognitive treatment is appropriate:

\begin{equation}
\begin{split}
&(\hat m_i,\hat r_i) = F_{\theta}(e_i), \\
&\theta^{*} = \arg\min_{\theta} \sum_{(e_i,m_i,r_i,y_i)\in\mathcal{D}_t} \left[ \lambda_m\mathcal{L}_{m}(\hat m_i,m_i) + \lambda_r\mathcal{L}_{r}(\hat r_i,r_i) \right].
\end{split}
\label{eq:continuous_learning}
\end{equation}

The first loss therefore learns

\begin{equation}
e_i \longrightarrow m_i,
\label{eq:event_to_treatment}
\end{equation}

namely, what cognitive method should be used for what kind of event.  The second
learns the result associated with the treated event.  This distinction prevents the
learning problem from collapsing into a direct event-to-action or event-to-outcome
mapping.  Two events may require different treatments even if their final actions are
similar, while different surface objects may share the same underlying cognitive
treatment.

As experience accumulates, a treatment that initially required explicit reasoning can
become automatic.  A first-time event may therefore follow

\[
e
\rightarrow
\mathrm{delib/fallback}
\rightarrow
m
\rightarrow
y
\rightarrow
\mathrm{learn},
\]

whereas a later sufficiently similar event can follow

\[
e'
\rightarrow
\mathrm{auto}
\rightarrow
m .
\]

Automatic processing is consequently not treated as the absence of cognition.  It is
a learned implementation of a cognitive treatment that was previously established
and validated through cognitive processing.

The environment may also change after a treatment has been learned.  Let
$\tilde r_t$ be the current prediction or cognitive result and let $y_{t+\Delta}$ be
a later validation signal.  Their consistency is evaluated by

\begin{equation}
v_t = V(\tilde r_t,y_{t+\Delta}).
\label{eq:delayed_validation}
\end{equation}

If the delayed evidence supports the previous treatment, the experience reinforces
the learned mechanism.  If it contradicts the previous result, the event becomes a
corrective learning material and may also reduce confidence in the corresponding
automatic processor, causing explicit cognition to be re-engaged.  Hence, previously
automatic cognition is not permanently fixed; it can be reopened, revised, and
subsequently re-automated when the environment changes.

Overall, the proposed cognitive cycle is

\begin{equation}
\boxed{
\begin{aligned}
\mathrm{encounter} & \rightarrow \mathrm{register} \rightarrow \mathrm{cover} \rightarrow \mathrm{select\ treatment}\\
&\rightarrow \mathrm{process}  \rightarrow \mathrm{validate} \rightarrow \mathrm{learn} \rightarrow \mathrm{reuse/revise}
\end{aligned}
}
\label{eq:overall_cognitive_cycle}
\end{equation}

The central objective is therefore not to maximize explicit reasoning.  It is to
maintain continuous cognitive coverage: every cognitively admitted event receives an
event-dependent treatment, familiar treatments are reused automatically when reliable,
and unfamiliar or changed events re-enter explicit cognition when necessary.  This
combination enables the robot to continuously process different encountered events in
different ways while continuously learning how such events should be treated.

\section{Verification}
\label{sec:verification}

This section evaluates three hypotheses that directly correspond to the proposed
model: (i) different encountered events should receive different cognitive
treatments according to their context rather than a fixed object-level rule;
(ii) cognitive processing should remain continuous when multiple events arrive,
overlap, or wait for delayed evidence; and (iii) the robot should continuously
learn, revise, and reuse event-dependent cognitive treatments. The experiments
use a controlled event-level mobile-service-robot simulator so that these
cognitive mechanisms can be isolated from low-level perception and motor-control
errors. All reported results are the mean and sample standard deviation over ten
independent seeds.

\subsection{Experimental Setup}
\label{subsec:verification_setup}

\paragraph{Robot scenario and event representation.}
The simulated robot continuously navigates and performs delivery, inspection, and
route-related activities in an indoor service/warehouse environment. During this
operation, it encounters objects and conditions such as normal or unstable objects,
blocked routes, changed object positions, abnormal devices, and previously unseen
objects. An admitted event contains the object type, location, state, recent-history
status, route relation, load information, and task/safety context. The experiments
therefore operate at the cognitive-event level: they test how an already registered
event is treated, scheduled, learned, and revised, rather than benchmarking visual
recognition itself.

The available cognitive treatments are \emph{description}, \emph{memory},
\emph{risk prediction}, \emph{planning}, \emph{diagnosis}, and \emph{analogy}.
Description performs lightweight semantic registration; memory is used when a
meaningful difference from historical state must be examined; risk prediction
evaluates unsafe future consequences; planning handles active route or task
constraints; diagnosis analyzes abnormal device states; and analogy provides a
general treatment for genuinely unfamiliar cases. These labels are sufficiently
different to test whether the robot learns an event-dependent treatment repertoire
rather than one universal reasoning procedure.

\paragraph{Models and implementation.}
Explicit deliberation uses \texttt{deepseek-v4-flash} through an
OpenAI-compatible interface with temperature $0$. In Experiment~1, automatic
treatment selection is implemented by a PyTorch multilayer perceptron with two
64-unit hidden layers, ReLU activations, dropout $0.1$, and a six-class output.
In Experiment~3, a shared two-layer 64-unit encoder feeds a six-class treatment
head and a two-class outcome head. The neural models use AdamW with learning rate
$10^{-3}$ and weight decay $10^{-4}$. Pretraining uses 140 epochs; continual
updates in Experiment~3 use 35 epochs. Experiment~2 intentionally does not invoke
the LLM or a neural learner because it isolates the continuous process-management
mechanism.

The experiment sizes and workload construction are summarized in
Table~\ref{tab:verification_config}. Training, calibration, and test object identities
in Experiment~1 are disjoint. For the proposed method, the automatic-to-deliberative
routing threshold is selected only on the calibration set from candidate values
$0.45$--$0.90$ and is frozen before testing. The bursty-delayed workload in
Experiment~2 deliberately combines clustered arrivals, more critical events, and
delayed evidence.

\begin{table*}[t]
\centering
\caption{Verification configuration. Event counts in Experiment~2 are the
mean generated counts over ten seeds; other counts are fixed by design.}
\label{tab:verification_config}
\scriptsize
\begin{tabular}{llll}
\toprule
Experiment & Training / initialization & Evaluation stream & Main stress factor \\
\midrule
Exp.~1 (V3) &
268 training + 42 calibration events &
82 test events (34 structured + 48 random) &
held-out object identities and context changes \\
Exp.~2 (V2) &
220 nominal simulator steps &
Low: 13.2; Medium: 43.7; High: 84.8; Bursty-delayed: 54.0 events &
concurrency, bursts, deadlines, and delayed evidence \\
Exp.~3 (V2) &
198 pretraining materials &
64 events (28 familiar + 6 novel + 24 drift + 6 reuse) &
novelty, delayed validation, outcome change, and reuse \\
\bottomrule
\end{tabular}
\end{table*}

\paragraph{Baselines.}
Experiment~1 compares the proposed calibrated hybrid mechanism with
\emph{Task-Only}, which treats only events considered relevant to the current task
or safety; \emph{Fixed Treatment}, which maps object categories to predefined
cognitive treatments without contextual adaptation; \emph{DL-Only}, which always
uses the learned treatment selector; and \emph{Always-LLM}, which explicitly
deliberates on every event. Experiment~2 uses \emph{Task-Bounded}, which ignores
events outside the active task; \emph{Sequential}, which must finish or wait on the
current cognitive process before proceeding; and \emph{Reactive Priority}, which
can preempt a process for a higher-priority event but does not preserve interrupted
cognitive state. Experiment~3 compares with \emph{Always-LLM};
\emph{Offline MultiHead}, which has the same treatment/outcome neural architecture
but does not update after deployment; and \emph{Continual Outcome-Only}, which
continually learns the final outcome but does not learn an event-dependent cognitive
treatment.

\paragraph{Metrics.}
Experiment~1 reports treatment accuracy, accuracy on the structured contextual
challenge, random-event robustness, task/safety accuracy, and the fraction of test
events processed automatically. Its strict context-pair success rate counts a pair
as correct only when both versions of the same contextual contrast are assigned the
correct treatments. Experiment~2 uses cognitive coverage,
\begin{equation}
C_{\mathrm{cov}} =
\frac{N_{\mathrm{completed\ cognitive\ events}}}
     {N_{\mathrm{admitted\ cognitive\ events}}},
\label{eq:coverage_metric}
\end{equation}
together with critical-event miss rate, critical-deadline success, treatment
latency, and primary-task completion. Experiment~3 uses treatment accuracy, outcome
accuracy, and joint accuracy, where a case is jointly correct only if both the
selected treatment and its outcome are correct. Novel-reuse automatic rate measures
whether a first-time explicitly processed event can later be handled automatically.
Delayed-revision success measures whether a changed event that is initially handled
incorrectly becomes correct after delayed validation and subsequent continual
updates.

\subsection{Experiment 1: Event-Dependent Cognitive Treatment}
\label{subsec:exp1_verification}

\paragraph{Purpose and scenario.}
The first experiment tests whether the robot can learn \emph{different ways of
thinking about different events}. A simple object-class mapping is insufficient for
this hypothesis, so the principal test uses held-out contextual contrasts. For
example, an object safely located at the center of a surface requires only
description, whereas the same type of object becomes a risk-prediction case when it
is unstable at an edge. Likewise, an object outside the active route can require
only description, but the same type of object blocking the active route requires
planning. Normal versus abnormally behaving devices and unchanged versus unexpectedly
moved objects create diagnosis and memory contrasts, respectively. The test objects
are not used in training, so successful treatment selection requires transfer of the
contextual principle rather than memorization of a specific object template.

\begin{table*}[t]
\centering
\caption{Experiment~1 results over ten seeds. Strict pair success is defined
over the 16 genuine two-event contextual contrasts; the two singleton
unknown-object cases are not pair tests.}
\label{tab:exp1_results}
\scriptsize
\resizebox{\textwidth}{!}{
\begin{tabular}{lccccccc}
\toprule
Method &
Overall treatment &
Structured &
Strict pair &
Random robustness &
Task/safety &
Automatic &
LLM calls \\
& accuracy & accuracy & success & accuracy & success & processing & per seed \\
\midrule
Task-Only & 23.41 $\pm$ 4.59 & 23.53 $\pm$ 0.00 & 0.00 $\pm$ 0.00 & 23.33 $\pm$ 7.84 & 49.23 $\pm$ 10.46 & 0.00 $\pm$ 0.00 & 0.0 $\pm$ 0.0 \\
Fixed Treatment & 31.59 $\pm$ 3.07 & 35.29 $\pm$ 0.00 & 0.00 $\pm$ 0.00 & 28.96 $\pm$ 5.24 & 26.24 $\pm$ 3.32 & 0.00 $\pm$ 0.00 & 0.0 $\pm$ 0.0 \\
DL-Only & 91.22 $\pm$ 3.09 & 94.41 $\pm$ 2.92 & 91.25 $\pm$ 3.23 & 88.96 $\pm$ 4.51 & 87.23 $\pm$ 6.87 & 100.00 $\pm$ 0.00 & 0.0 $\pm$ 0.0 \\
Always-LLM & 66.34 $\pm$ 3.95 & 97.35 $\pm$ 0.93 & 100.00 $\pm$ 0.00 & 44.38 $\pm$ 6.88 & 59.04 $\pm$ 6.79 & 0.00 $\pm$ 0.00 & 82.0 $\pm$ 0.0 \\
Proposed & 92.44 $\pm$ 3.29 & 96.76 $\pm$ 1.67 & 96.25 $\pm$ 3.23 & 89.38 $\pm$ 5.51 & 88.68 $\pm$ 8.22 & 93.66 $\pm$ 4.33 & 5.2 $\pm$ 3.6 \\
\bottomrule
\end{tabular}}
\end{table*}

Table~\ref{tab:exp1_results} shows that a fixed object-level treatment is inadequate:
Fixed Treatment reaches only $35.29\%$ structured accuracy and fails all strict
context pairs, while Task-Only reaches $23.53\%$ structured accuracy. DL-Only learns
much of the contextual mapping, obtaining $94.41\pm2.92\%$ structured accuracy and
$91.25\pm3.23\%$ strict pair success. The proposed method further increases these
results to $96.76\pm1.67\%$ and $96.25\pm3.23\%$, respectively. This difference is
important because the held-out pairs change the context while preserving the
underlying object role; the gain therefore reflects improved selection of
\emph{how the event should be cognitively treated}, rather than memorization of
which object is being observed.

Always-LLM is strongest on the deliberately unambiguous structured pairs, reaching
$97.35\pm0.93\%$ structured accuracy and $100\%$ strict pair success, but it invokes
the LLM for all 82 test events. In contrast, the proposed method processes
$93.66\pm4.33\%$ of the test events automatically and requires only
$5.2\pm3.6$ LLM calls per seed. Thus, it retains nearly the same performance as
full deliberation on the structured cognitive challenge while replacing most
repeated reasoning with learned processing. On the randomly mixed robustness cases,
the proposed method obtains $89.38\pm5.51\%$, comparable to DL-Only's
$88.96\pm4.51\%$, showing that calibrated fallback preserves the learned
automatic competence on mixed cases while improving the structured contextual
challenge.

\begin{figure}[t]
\centering
\includegraphics[width=\linewidth]{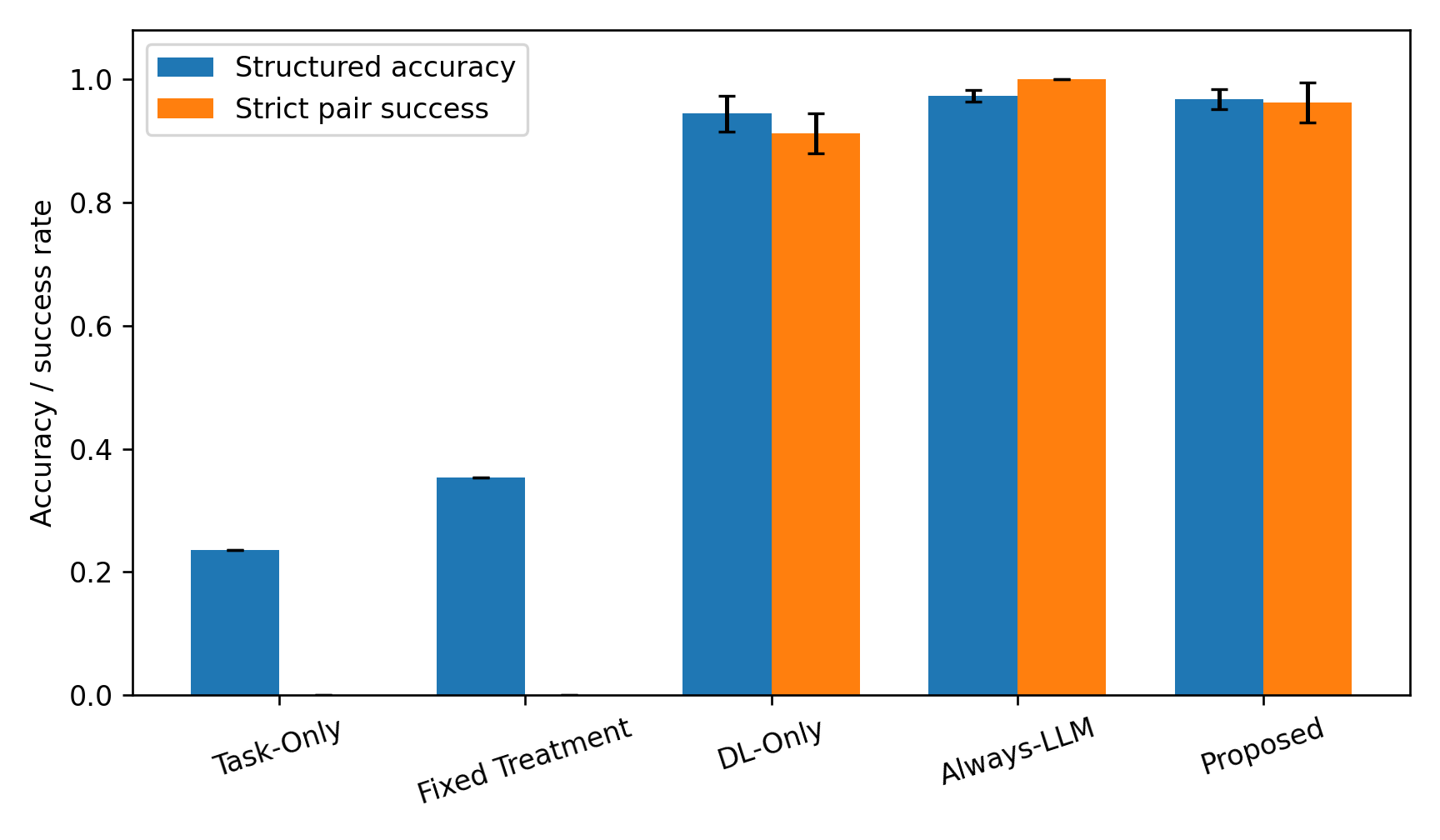}
\caption{Experiment~1: structured contextual-treatment accuracy and strict
context-pair success. The latter requires both members of a contextual contrast
to be treated correctly.}
\label{fig:exp1_context}
\end{figure}

Figure~\ref{fig:exp1_context} emphasizes the intended distinction. Fixed rules cannot
adapt when the same class of situation changes cognitive meaning, whereas the learned
selector captures most such changes and calibrated deliberation recovers additional
low-confidence cases. The experiment therefore supports the first hypothesis:
cognitive processing should be event-dependent rather than a single fixed procedure.

\subsection{Experiment 2: Continuous Cognitive Processing}
\label{subsec:exp2_verification}

\paragraph{Purpose and scenario.}
Experiment~2 tests whether cognitive treatment remains continuous when events overlap
in time. The robot executes its normal task while independent cognitive events arrive.
Some events require multiple processing steps, some are critical, and some cannot be
completed until delayed evidence appears. A delayed process may therefore need to be
suspended while another event is handled and later resumed. This distinction is
minor in a sparse environment but becomes important when cognitive workload exceeds
the rate at which a strictly sequential reasoner can finish it.

\begin{table}[t]
\centering
\caption{Experiment~2 cognitive coverage under increasing event load.}
\label{tab:exp2_coverage}
\scriptsize
\resizebox{\columnwidth}{!}{
\begin{tabular}{lcccc}
\toprule
Method & Low & Medium & High & Bursty-delayed \\
\midrule
Task-Bounded & 39.62 $\pm$ 12.75 & 34.40 $\pm$ 6.97 & 41.23 $\pm$ 7.22 & 37.05 $\pm$ 8.07 \\
Sequential & 100.00 $\pm$ 0.00 & 86.42 $\pm$ 13.00 & 40.43 $\pm$ 3.56 & 53.68 $\pm$ 13.61 \\
Reactive Priority & 82.67 $\pm$ 12.99 & 59.28 $\pm$ 4.23 & 46.89 $\pm$ 3.99 & 35.75 $\pm$ 5.64 \\
Proposed & 100.00 $\pm$ 0.00 & 99.45 $\pm$ 1.23 & 67.13 $\pm$ 7.94 & 92.64 $\pm$ 7.58 \\
\bottomrule
\end{tabular}}
\end{table}

\begin{table*}[t]
\centering
\caption{Experiment~2 stress-condition results. Miss and deadline columns refer
to critical events; latency is measured in simulator steps.}
\label{tab:exp2_stress}
\scriptsize
\resizebox{\textwidth}{!}{
\begin{tabular}{lcccccccc}
\toprule
& \multicolumn{4}{c}{High load} &
\multicolumn{4}{c}{Bursty-delayed} \\
\cmidrule(lr){2-5}\cmidrule(lr){6-9}
Method &
Critical miss & Critical deadline & Latency & Primary task &
Critical miss & Critical deadline & Latency & Primary task \\
\midrule
Task-Bounded & 54.60 $\pm$ 9.90 & 27.68 $\pm$ 9.20 & 9.05 $\pm$ 5.79 & 67.85 $\pm$ 7.62 & 65.93 $\pm$ 13.52 & 28.46 $\pm$ 11.47 & 2.90 $\pm$ 1.80 & 80.85 $\pm$ 8.08 \\
Sequential & 60.08 $\pm$ 6.92 & 0.43 $\pm$ 1.37 & 92.84 $\pm$ 15.06 & 45.70 $\pm$ 7.12 & 42.37 $\pm$ 17.24 & 4.67 $\pm$ 6.01 & 74.08 $\pm$ 19.00 & 42.22 $\pm$ 7.22 \\
Reactive Priority & 35.39 $\pm$ 8.18 & 62.31 $\pm$ 8.20 & 38.78 $\pm$ 9.69 & 49.40 $\pm$ 3.62 & 48.54 $\pm$ 13.16 & 50.66 $\pm$ 14.14 & 5.52 $\pm$ 4.04 & 61.35 $\pm$ 7.03 \\
Proposed & 0.00 $\pm$ 0.00 & 70.39 $\pm$ 11.36 & 40.53 $\pm$ 12.03 & 49.70 $\pm$ 5.37 & 0.00 $\pm$ 0.00 & 66.79 $\pm$ 16.62 & 26.66 $\pm$ 14.37 & 55.82 $\pm$ 5.57 \\
\bottomrule
\end{tabular}}
\end{table*}

At low load, both Proposed and Sequential achieve $100\%$ cognitive coverage
(Table~\ref{tab:exp2_coverage}); hence the proposed process manager is not needed to
manufacture an advantage in an easy setting. As the event stream becomes denser,
the difference grows. At medium load, Proposed maintains
$99.45\pm1.23\%$ coverage compared with Sequential's
$86.42\pm13.00\%$. Under high load, Proposed retains
$67.13\pm7.94\%$ coverage whereas Sequential falls to
$40.43\pm3.56\%$. The bursty-delayed setting is especially diagnostic:
Proposed completes $92.64\pm7.58\%$ of admitted events, compared with
$53.68\pm13.61\%$ for Sequential, $35.75\pm5.64\%$ for Reactive Priority,
and $37.05\pm8.07\%$ for Task-Bounded.

The safety-related metrics explain why this higher coverage matters. In the
high-load condition, Proposed has a $0\%$ critical-event miss rate and achieves
$70.39\pm11.36\%$ critical-deadline success. Sequential misses
$60.08\pm6.92\%$ of critical events and has only
$0.43\pm1.37\%$ critical-deadline success. The corresponding treatment latency is
$40.53\pm12.03$ steps for Proposed versus $92.84\pm15.06$ for Sequential.
In the bursty-delayed condition, Proposed again has $0\%$ critical misses and
$66.79\pm16.62\%$ critical-deadline success, whereas Sequential misses
$42.37\pm17.24\%$ of critical events and reaches only
$4.67\pm6.01\%$ critical-deadline success.

Task-Bounded preserves more primary-task progress in the bursty-delayed case
($80.85\pm8.08\%$ versus $55.82\pm5.57\%$ for Proposed), but
Table~\ref{tab:exp2_coverage} shows why: it cognitively covers only
$37.05\pm8.07\%$ of admitted events and misses
$65.93\pm13.52\%$ of critical events. Thus, concentrating only on the current
explicit task is efficient precisely because much of the surrounding cognitive
workload is discarded. The proposed method instead maintains broad cognitive
coverage while preserving interrupted process state.

\begin{figure}[t]
\centering
\includegraphics[width=\linewidth]{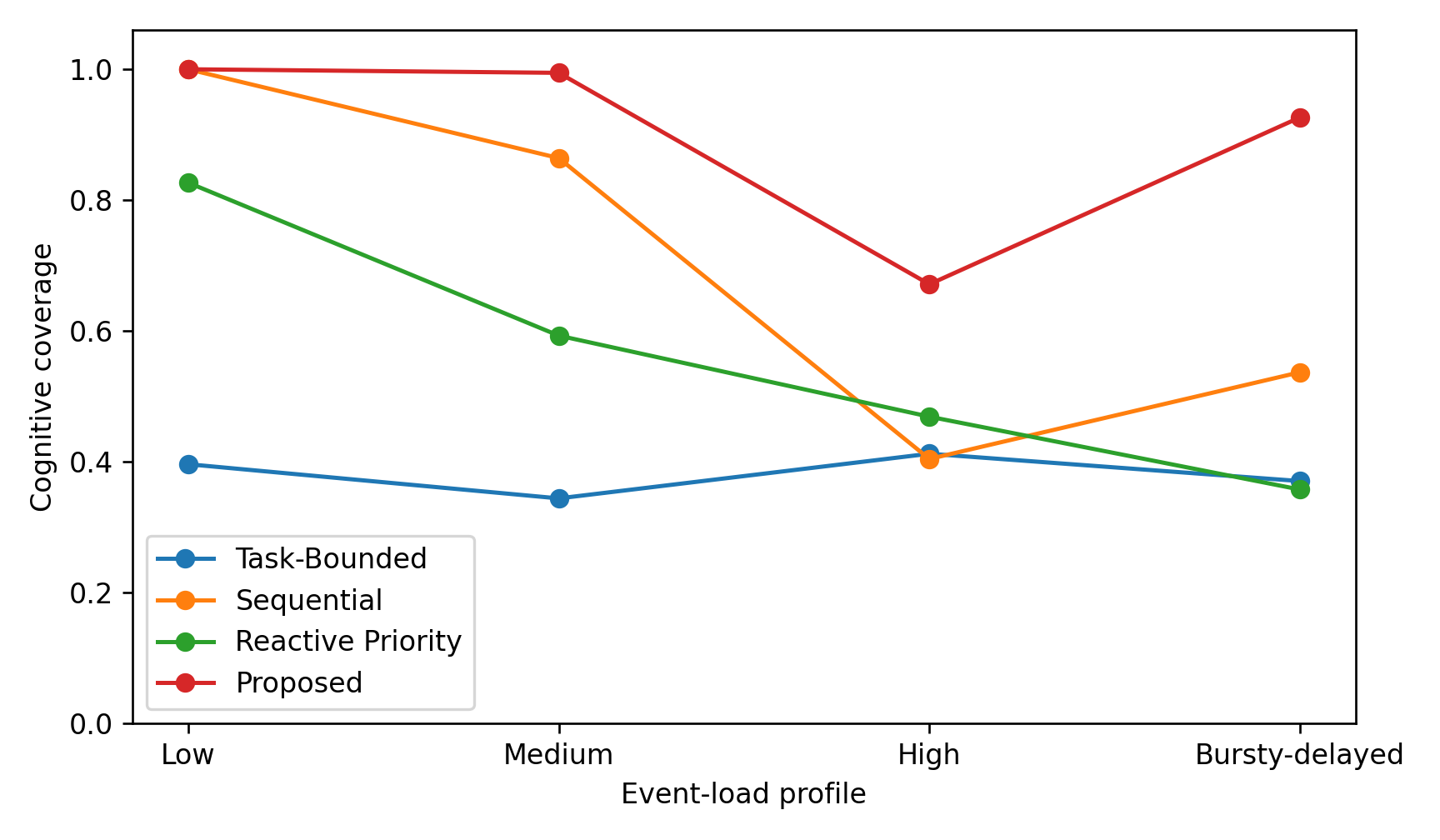}
\caption{Experiment~2: cognitive coverage as the event workload becomes more
demanding. The advantage of continuous suspend/resume processing emerges as
events become dense or bursty and delayed.}
\label{fig:exp2_coverage}
\end{figure}

Figure~\ref{fig:exp2_coverage} shows the central trend: the methods are closer under
light workload, but the gap widens as cognition becomes concurrent. Reactive Priority
can respond rapidly to new urgent events, yet its lack of preserved interrupted
state limits total coverage. Sequential processing preserves state but becomes
blocked by unresolved events. The proposed mechanism combines preemption with
state-preserving suspension and resumption, supporting the second hypothesis that
continuous cognition requires management of multiple incomplete cognitive processes.

\subsection{Experiment 3: Continuous Learning, Revision, and Reuse}
\label{subsec:exp3_verification}

\paragraph{Purpose and scenario.}
The third experiment tests whether learning is itself continuous and event-dependent.
The robot first encounters familiar events, then six previously unseen event types,
followed by a controlled change in the environment where four previously learned
observable event signatures produce different outcomes. Finally, the six novel
events appear again. This stream tests three requirements simultaneously: a new event
should initially receive explicit/fallback treatment; validated experience should
later make that treatment reusable; and previously reliable automatic cognition
should be revised when delayed outcomes reveal that the environment has changed.

\begin{table}[htpb]
\centering
\caption{Experiment~3 overall continual-learning results.}
\label{tab:exp3_overall}
\scriptsize
\resizebox{\columnwidth}{!}{
\begin{tabular}{lccccc}
\toprule
Method & Treatment & Outcome & Joint & Automatic & LLM calls \\
& accuracy & accuracy & accuracy & processing & per seed \\
\midrule
Always-LLM & 74.84 $\pm$ 3.86 & 76.09 $\pm$ 1.96 & 60.00 $\pm$ 3.39 & 0.00 $\pm$ 0.00 & 64.0 $\pm$ 0.0 \\
Offline MultiHead & 86.56 $\pm$ 2.68 & 63.28 $\pm$ 5.12 & 53.91 $\pm$ 5.27 & 53.12 $\pm$ 6.79 & 30.0 $\pm$ 4.3 \\
Continual Outcome-Only & 43.12 $\pm$ 2.78 & 82.66 $\pm$ 2.70 & 34.06 $\pm$ 2.83 & 88.12 $\pm$ 1.68 & 7.6 $\pm$ 1.1 \\
Proposed & 92.34 $\pm$ 2.01 & 85.78 $\pm$ 2.14 & 79.53 $\pm$ 3.16 & 75.62 $\pm$ 2.35 & 15.6 $\pm$ 1.5 \\
\bottomrule
\end{tabular}}
\end{table}

\begin{table*}[htpb]
\centering
\caption{Experiment~3 adaptation to novel events and changed outcomes.}
\label{tab:exp3_adaptation}
\scriptsize
\resizebox{\textwidth}{!}{
\begin{tabular}{lcccccc}
\toprule
Method &
Novel first joint &
Novel reuse joint &
Novel reuse automatic &
Drift initial joint &
Drift late joint &
Delayed revision \\
\midrule
Always-LLM & 51.67 $\pm$ 5.27 & 83.33 $\pm$ 0.00 & 0.00 $\pm$ 0.00 & 25.00 $\pm$ 0.00 & 74.17 $\pm$ 2.64 & 66.67 $\pm$ 0.00 \\
Offline MultiHead & 50.00 $\pm$ 0.00 & 83.33 $\pm$ 0.00 & 0.00 $\pm$ 0.00 & 17.50 $\pm$ 12.08 & 17.50 $\pm$ 12.08 & 0.00 $\pm$ 0.00 \\
Continual Outcome-Only & 41.67 $\pm$ 8.78 & 83.33 $\pm$ 0.00 & 100.00 $\pm$ 0.00 & 25.00 $\pm$ 0.00 & 50.00 $\pm$ 0.00 & 33.33 $\pm$ 0.00 \\
Proposed & 50.00 $\pm$ 0.00 & 100.00 $\pm$ 0.00 & 100.00 $\pm$ 0.00 & 25.00 $\pm$ 0.00 & 99.17 $\pm$ 2.64 & 100.00 $\pm$ 0.00 \\
\bottomrule
\end{tabular}}
\end{table*}

The overall results in Table~\ref{tab:exp3_overall} show that continuously learning
both the cognitive treatment and its result is important. Proposed obtains
$92.34\pm2.01\%$ treatment accuracy, $85.78\pm2.14\%$ outcome accuracy, and
$79.53\pm3.16\%$ joint accuracy. Offline MultiHead reaches only
$53.91\pm5.27\%$ joint accuracy because it cannot incorporate the newly validated
experience. Continual Outcome-Only attains a relatively high
$82.66\pm2.70\%$ outcome accuracy, but its treatment accuracy is only
$43.13\pm2.78\%$, reducing joint accuracy to $34.06\pm2.83\%$. Thus,
continuously learning the final result alone does not reproduce the proposed
event-dependent cognition; the robot must also learn \emph{which cognitive method}
is appropriate.

The proposed method also avoids reasoning from scratch on every event. It processes
$75.63\pm2.35\%$ of the stream automatically and uses
$15.6\pm1.5$ LLM calls per seed, compared with 64 calls for Always-LLM. More
importantly, Table~\ref{tab:exp3_adaptation} shows the intended transition from
unfamiliar to learned treatment. First-time novel events have
$50.00\%$ joint accuracy. After validation, their later occurrences reach
$100\%$ joint accuracy and $100\%$ automatic processing. This directly demonstrates the
reason--validate--learn--reuse path.

The changed-outcome block tests whether previously learned automatic cognition can be
corrected rather than becoming permanently fixed. Proposed begins this block at only
$25.00\%$ joint accuracy because the old learned outcomes are intentionally no longer
valid. After continual validation and updates, late drift accuracy rises to
$99.17\pm2.64\%$, and delayed-revision success reaches $100\%$.
Offline MultiHead remains at $17.50\pm12.08\%$ late drift accuracy with
$0\%$ revision success, while Continual Outcome-Only reaches
$50.00\%$ late drift accuracy and $33.33\%$ revision success. These results show
that continuous learning is required not only to acquire new treatments but also to
repair previously automated cognition when later evidence contradicts it.

\begin{figure}[t]
\centering
\includegraphics[width=\linewidth]{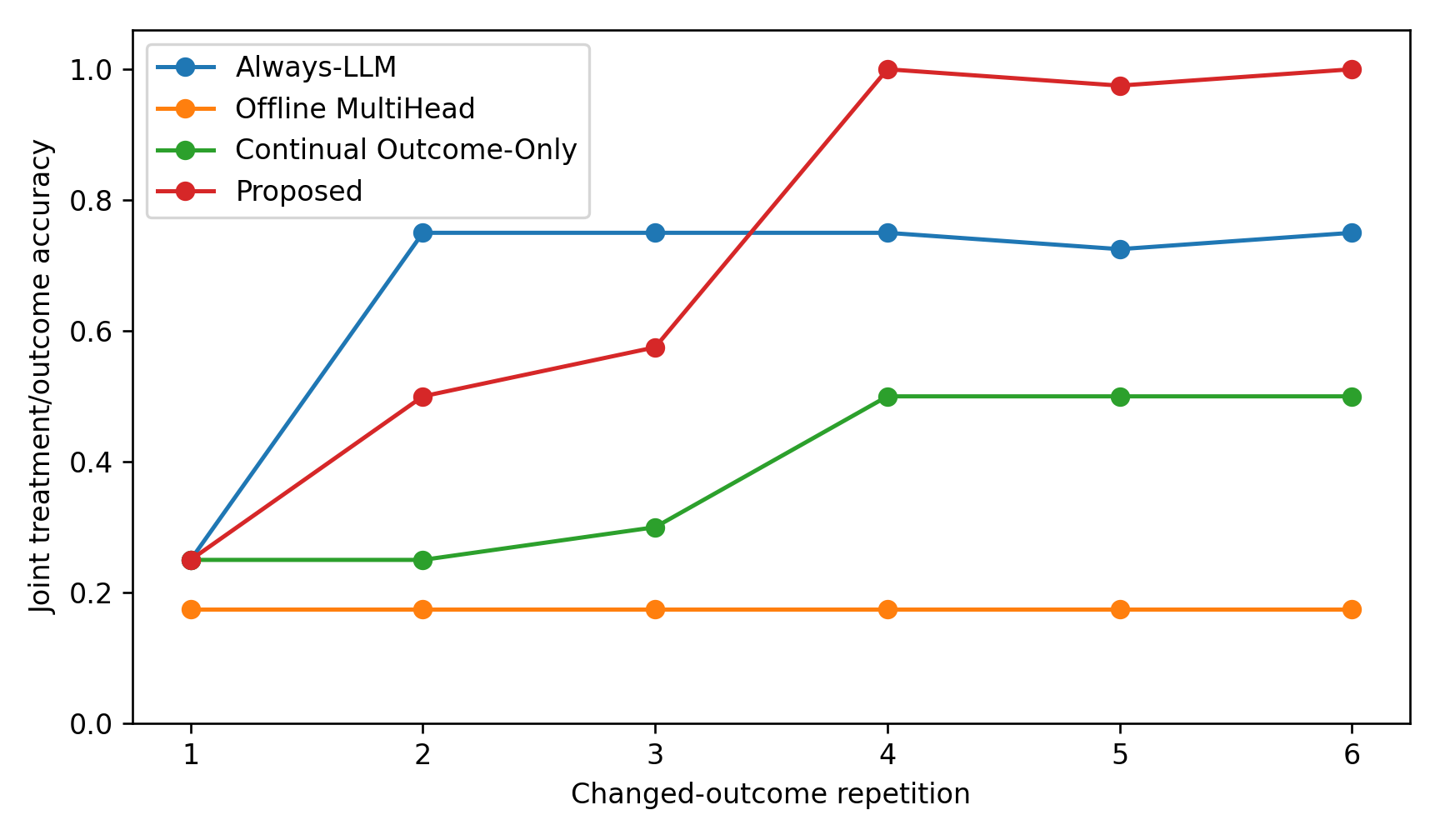}
\caption{Experiment~3: joint accuracy over six repetitions after the controlled
outcome change. Each repetition aggregates the four changed event signatures over
ten seeds.}
\label{fig:exp3_drift}
\end{figure}

As shown in Figure~\ref{fig:exp3_drift}, Proposed recovers progressively after the
environmental change: the average joint accuracy over the four drift signatures rises
from $25.0\%$ on the first occurrence to $50.0\%$, $57.5\%$, and $100\%$ by the
fourth occurrence, and remains at $97.5$--$100\%$ thereafter. In contrast, the
offline learner remains unchanged. This result completes the third hypothesis:
learning is a continuing cognitive process that can acquire different treatments for
different events, convert validated treatments into automatic processing, and reopen
previously learned cognition for revision when the environment changes.

\subsection{Summary of Findings}
\label{subsec:verification_summary}

The three experiments test complementary parts of the model rather than repeating the
same task-level success measure. Experiment~1 shows that the robot can select
context-dependent cognitive treatments and retain near-full-deliberation performance
while automatically handling most events. Experiment~2 shows why cognition must remain
continuous: suspend/resume state management becomes advantageous only when multiple
events overlap or await delayed evidence. Experiment~3 shows that the treatment
repertoire is not static; treatments are continuously learned, automatically reused,
and revised when their previous outcomes cease to hold. Together, the results support
the proposed view that autonomous cognition should consist of continuous,
event-dependent processing coupled with continuous event-dependent learning.

\section{Conclusion}
\label{sec:conclusion}

This paper presented a continuous cognitive coverage framework for autonomous robots, in which every cognitively admitted event receives an appropriate event-dependent cognitive treatment rather than being processed only when an explicit task demands reasoning. The framework distinguishes what cognitive treatment should be applied from how it should be executed, allowing familiar events to be handled automatically while unfamiliar, uncertain, or changed events invoke explicit deliberation or fallback reasoning. Multiple cognitive processes can be suspended, resumed, and interleaved to maintain continuous cognition, while validated experiences are continually used to learn, automate, and revise event-specific treatments.

Experimental results support these mechanisms. The proposed method achieved $96.76\%$ structured treatment accuracy while automatically processing $93.66\%$ of events, maintained $92.64\%$ cognitive coverage under bursty-delayed workloads, and reached $79.53\%$ joint accuracy in continual learning. Novel events were subsequently reused with $100\%$ automatic processing, while changed outcomes were successfully incorporated through continual revision. These results indicate that continuous cognitive coverage can provide a practical basis for robots that continuously process different encountered events in different ways while progressively learning how those events should be treated.

% \section*{Acknowledgment}
% The authors thanks th.

\ifCLASSOPTIONcaptionsoff
  \newpage
\fi

\bibliographystyle{IEEEtran}
\bibliography{ref}

@article{franklin2012lida,
  title={Global workspace theory, its LIDA model and the underlying neuroscience},
  author={Franklin, Stan and Strain, Steve and Snaider, Javier and McCall, Ryan and Faghihi, Usef},
  journal={Biologically Inspired Cognitive Architectures},
  volume={1},
  pages={32--43},
  year={2012},
  publisher={Elsevier}
}

@article{helie2014autonomous,
  title={Autonomous learning in psychologically-oriented cognitive architectures: A survey},
  author={H{\'e}lie, S{\'e}bastien and Sun, Ron},
  journal={New Ideas in Psychology},
  volume={34},
  pages={37--55},
  year={2014},
  publisher={Elsevier}
}

@article{sun2001implicit,
  title={From implicit skills to explicit knowledge: A bottom-up model of skill learning},
  author={Sun, Ron and Merrill, Edward and Peterson, Todd},
  journal={Cognitive science},
  volume={25},
  number={2},
  pages={203--244},
  year={2001},
  publisher={Wiley Online Library}
}

@article{ichter2023saycan,
  title={Do as i can, not as i say: Grounding language in robotic affordances},
  author={Ahn, Michael and Brohan, Anthony and Brown, Noah and Chebotar, Yevgen and Cortes, Omar and David, Byron and Finn, Chelsea and Fu, Chuyuan and Gopalakrishnan, Keerthana and Hausman, Karol and others},
  journal={arXiv preprint arXiv:2204.01691},
  year={2022}
}

@article{huang2023inner,
  title={Inner monologue: Embodied reasoning through planning with language models},
  author={Huang, Wenlong and Xia, Fei and Xiao, Ted and Chan, Harris and Liang, Jacky and Florence, Pete and Zeng, Andy and Tompson, Jonathan and Mordatch, Igor and Chebotar, Yevgen and others},
  journal={arXiv preprint arXiv:2207.05608},
  year={2022}
}

@article{driess2023palme,
  title={Palm-e: An embodied multimodal language model},
  author={Driess, Danny and Xia, Fei and Sajjadi, Mehdi SM and Lynch, Corey and Chowdhery, Aakanksha and Ichter, Brian and Wahid, Ayzaan and Tompson, Jonathan and Vuong, Quan and Yu, Tianhe and others},
  journal={arXiv preprint arXiv:2303.03378},
  year={2023}
}

@article{zitkovich2023rt2,
  title={Rt-2: Vision-language-action models transfer web knowledge to robotic control},
  author={Brohan, Anthony and Brown, Noah and Carbajal, Justice and Chebotar, Yevgen and Chen, Xi and Choromanski, Krzysztof and Ding, Tianli and Driess, Danny and Dubey, Avinava and Finn, Chelsea and others},
  journal={arXiv preprint arXiv:2307.15818},
  year={2023}
}

@article{yao2023react,
  title={React: Synergizing reasoning and acting in language models},
  author={Yao, Shunyu and Zhao, Jeffrey and Yu, Dian and Du, Nan and Shafran, Izhak and Narasimhan, Karthik and Cao, Yuan},
  journal={arXiv preprint arXiv:2210.03629},
  year={2022}
}

@article{lesort2020continual,
  title={Continual learning for robotics: Definition, framework, learning strategies, opportunities and challenges},
  author={Lesort, Timoth{\'e}e and Lomonaco, Vincenzo and Stoian, Andrei and Maltoni, Davide and Filliat, David and D{\'\i}az-Rodr{\'\i}guez, Natalia},
  journal={Information fusion},
  volume={58},
  pages={52--68},
  year={2020},
  publisher={Elsevier}
}

@article{wang2024voyager,
  title={Voyager: An open-ended embodied agent with large language models},
  author={Wang, Guanzhi and Xie, Yuqi and Jiang, Yunfan and Mandlekar, Ajay and Xiao, Chaowei and Zhu, Yuke and Fan, Linxi and Anandkumar, Anima},
  journal={arXiv preprint arXiv:2305.16291},
  year={2023}
}

@article{shinn2023reflexion,
  title={Reflexion: Language agents with verbal reinforcement learning},
  author={Shinn, Noah and Cassano, Federico and Gopinath, Ashwin and Narasimhan, Karthik and Yao, Shunyu},
  journal={Advances in neural information processing systems},
  volume={36},
  pages={8634--8652},
  year={2023}
}

@article{yu2024distilling,
  title={Distilling system 2 into system 1},
  author={Yu, Ping and Xu, Jing and Weston, Jason and Kulikov, Ilia},
  journal={arXiv preprint arXiv:2407.06023},
  year={2024}
}


\begin{IEEEbiography}{Hong Su}
  received the MS and PhD degrees, in 2006 and 2022, respectively, from Sichuan University, Chengdu, China. He is currently a researcher of Chengdu University of Information Technology Chengdu, China. His research interests include blockchain, large language model and human simulation computing.
\end{IEEEbiography}


%\vfill


\end{document}